\documentclass[sigconf]{acmart}

\AtBeginDocument{%
  \providecommand\BibTeX{{%
    \normalfont B\kern-0.5em{\scshape i\kern-0.25em b}\kern-0.8em\TeX}}}

\makeatletter
\def\@ACM@checkaffil{
    \if@ACM@instpresent\else
    \ClassWarningNoLine{\@classname}{No institution present for an affiliation}%
    \fi
    \if@ACM@citypresent\else
    \ClassWarningNoLine{\@classname}{No city present for an affiliation}%
    \fi
    \if@ACM@countrypresent\else
        \ClassWarningNoLine{\@classname}{No country present for an affiliation}%
    \fi
}
\makeatother

\acmConference[ICCAD '23]{ICCAD '23: Fast Machine Learning for Science Workshop}{October 29--November 2, 2023}{CA, USA}
\acmBooktitle{ICCAD '23: Fast Machine Learning for Science Workshop, October 29--November 2, 2023, CA, USA}

\usepackage[utf8]{inputenc} 
\usepackage[T1]{fontenc}    
\usepackage{hyperref}       
\usepackage{url}            
\usepackage{booktabs}       
\usepackage{amsfonts}       
\usepackage{nicefrac}       
\usepackage{microtype}      

\usepackage{amsmath}
\usepackage{algorithm}
\usepackage{algorithmic}
\usepackage{graphicx,dblfloatfix,caption,afterpage}
\usepackage{textcomp}
\usepackage[scaled=0.95]{couriers}
\usepackage{subfigure}
\usepackage{multirow}
\usepackage{makecell}
\usepackage{mathtools, nccmath}
\usepackage{array}
\usepackage{multirow} 
\usepackage{soul}
\usepackage{siunitx}
\usepackage{colortbl}
\usepackage{arydshln}
\usepackage{booktabs}
\usepackage{balance}

\graphicspath{{figures/}}

\begin{document}

\title{Toward Reinforcement Learning–based Rectilinear Macro Placement Under Human Constraints}

\author{Tuyen P. Le}
\email{tuyenple@agilesoda.ai}
\affiliation{%
  \institution{AgileSoDA Company}
  \city{Seoul}
  \state{South Korea}
  \postcode{06149}
}
\author{Hieu T. Nguyen}
\email{trong@agilesoda.ai}
\affiliation{%
  \institution{AgileSoDA Company}
  \city{Seoul}
  \state{South Korea}
  \postcode{06149}
}
\author{Seungyeol Baek}
\email{sybaek@agilesoda.ai}
\affiliation{%
  \institution{AgileSoDA Company}
  \city{Seoul}
  \state{South Korea}
  \postcode{06149}
}
\author{Taeyoun Kim}
\email{klumblr@agilesoda.ai}
\affiliation{%
  \institution{AgileSoDA Company}
  \city{Seoul}
  \state{South Korea}
  \postcode{06149}
}
\author{Jungwoo Lee}
\email{justinlee@agilesoda.ai}
\affiliation{%
  \institution{AgileSoDA Company}
  \city{Seoul}
  \state{South Korea}
  \postcode{06149}
}
\author{Seongjung Kim}
\email{seongjung@asicland.com}
\affiliation{%
  \institution{Asicland Company}
  \city{Suwon}
  \state{South Korea}
  \postcode{06149}
}
\author{Hyunjin Kim}
\email{tkv93@asicland.com}
\affiliation{%
  \institution{Asicland Company}
  \city{Suwon}
  \state{South Korea}
  \postcode{06149}
}
\author{Misu Jung}
\email{msjung92@asicland.com}
\affiliation{%
  \institution{Asicland Company}
  \city{Suwon}
  \state{South Korea}
  \postcode{06149}
}
\author{Daehoon Kim}
\email{dhkim@asicland.com}
\affiliation{%
  \institution{Asicland Company}
  \city{Suwon}
  \state{South Korea}
  \postcode{06149}
}
\author{Seokyong Lee}
\email{sean.lee@asicland.com}
\affiliation{%
  \institution{Asicland Company}
  \city{Suwon}
  \state{South Korea}
  \postcode{06149}
}
\author{Daewoo Choi}
\email{daewoo.choi@hufs.ac.kr}
\affiliation{%
  \institution{Hankuk University of Foreign Studies}
  \city{Yongin}
  \state{South Korea}
  \postcode{06149}
}
\renewcommand{\shortauthors}{Tuyen P. Le, et al.}

\begin{abstract}
Macro placement is a critical phase in chip design, which becomes more intricate when involving general rectilinear macros and layout areas. Furthermore, macro placement that incorporates human-like constraints, such as design hierarchy and peripheral bias, has the potential to significantly reduce the amount of additional manual labor required from designers. This study proposes a methodology that leverages an approach suggested by Google's Circuit Training (G-CT) to provide a learning-based macro placer that not only supports placing rectilinear cases, but also adheres to crucial human-like design principles. Our experimental results demonstrate the effectiveness of our framework in achieving power-performance-area (PPA) metrics and in obtaining placements of high quality, comparable to those produced with human intervention. Additionally, our methodology shows potential as a generalized model to address diverse macro shapes and layout areas.
\end{abstract}



\maketitle

\section{Introduction}
Placement of macros is a vital and time-consuming process in chip design and substantially affects the power, performance, and area (PPA) metrics. The increasing demand for customized ASIC chip designs has complicated this task, especially for rectilinear, i.e., non-rectangular,  chip layouts and macro shapes. Moreover, development of macro placement algorithms that allow various design constraints, including human-like rules, is challenging and time-consuming.

Classical methods, such as partitioning-based methods and heuristic methods, were widely adopted in macro placers due to their straightforwardness and stability. However, they become time-consuming when dealing with advanced technology nodes and large-scale netlists. Modern macro placers using analytical methods have shown efficiency in representing highly complex objective functions and handling large-scale netlists. However, they lack insights for dealing with the ever-growing diversity of chip designs, continually restarting the process rather than building on prior lessons. Learning-based methods can advance this process because the model learned from past designs so it can predict better placements for unseen designs. Specially, a recent reinforcement learning (RL)-based macro placer proposed by Google \cite{mirhoseini2021graph} has shown potential as a generalized placer to rapidly predict high-quality macro placements for new, unseen designs.

Our methodology, leveraged circuit training (A-CT)-based macro placement, represents an advance to the learning-based approach, but still enjoys advantages from both conventional approaches. Particularly, our placer uses a clustered netlist generated by a grouping engine that uses partitioning-based methods to reduce a large netlist and to maintain some of design hierarchy inherent in the netlist. The placer uses RL to train a deep neural network (the agent) to predict placements for unseen designs and to provide flexibility in dealing with rectilinear layout areas and macros. The placer output is strengthened by a simulated annealing (SA)-based placement engine, which aims for refinements to achieve human-like placement quality by satisfying additional constraints. Finally, our RL-based placement engine trains the agent to maximize the total discounted reward function, which serves as a multiple objective function that integrates various analytical metrics in chip design. Our key contributions are as follows.

\begin{itemize}

\item We propose enhancements to CT-based macro placement including a core RL-based engine that is supported by conventional methods for grouping the netlist, and fine-tuning placement to account for human-like constraints (such as placing macros based on design hierarchy guided from the netlist, placing macros at the periphery, and pin accessibility constraints).

\item We present methods to unify macro placement using macros and layout areas for general rectilinear shapes. To the best of our knowledge, this is the first work dealing with rectilinear layout areas and macro shapes using RL.

\item We propose an enhanced RL model and demonstrate that our RL-based placer can use fewer resources than previously reported and still achieve competitive PPA metrics both from proxies and standard commercial tools.

\end{itemize}

\section{Preliminary}

\subsection{Review of Previous Work}

\textbf{Non-rectangular macro and areas placement.} Macros can have arbitrary shapes, and many studies \cite{10.1145/288548.288623, 10.1109/72.623207, 10.5555/244522.244865, 10.1109/43.144848} have proposed algorithms to place rectilinear macros, a special kind of non-rectangular macros with only 90 degrees angles. Given the increasing interest and investment in quantum computing, non-rectangular macro placement is likely to continue to be an important area of research in chip layout\cite{quantum}.\\
\textbf{Methods for macro placement.} Macro placement methods are classified into three categories: partitioning-based methods, heuristic methods, and analytical methods. Partitioning-based methods \cite{10.1145/1055137.1055184, 10.1145/378239.379064} use a divide-and-conquer strategy to recursively divide the chip areas and netlist into smaller sub-regions and sub-netlists and then assign each sub-netlist to a sub-region via min-cut objective functions. Partitioning-based methods are scalable, but the min-cut objective function has the drawback that it does not take explicitly into account common performance metrics such as wirelength, density, and congestion. Heuristic methods \cite{10.1109/ICCAD.2000.896483, 10.1109/ICISCE.2016.45} can consider performance metrics in their optimization functions and potentially reach a good placement. However, they are time-consuming and struggle to deal with very large circuit netlists. Analytical methods \cite{10.1109/TCAD.2008.923063, 10.1145/2699873, 10.1109/TCAD.2018.2859220} model the placement problem using mathematical techniques and use optimization methods to improve an objective function. Modern analytical methods can be efficient and scalable via parallelization on multi-threaded CPUs \cite{10.1109/TCAD.2018.2859220}, and by utilizing multiple GPUs \cite{10.1145/3316781.3317803}. In addition to these three classical categories, learning-based methods \cite{mirhoseini2021graph, lai2022maskplace, cheng2022the, cheng2021on, yang2022versatile} have been an active topic in academic research in recent years because of their potential to create a generalized placement model “averaged” from many designs. The Google method\cite{mirhoseini2021graph} trains a graph neural network (GNN) agent using an RL algorithm to place macros, and the trained agent has been shown to adapt well to unseen designs.\\
\textbf{Human-like constraints.} Physical designers consider various design features to produce high-quality placements. Some placers \cite{10.1145/3505170.3506731, 10.1145/3508352.3561379, 10.23919/DATE.2019.8714812} locate macros following the design hierarchy derived from the RTL model or directly from the cell-level netlist. Other placers locate macros near the periphery of the area to minimize the effects of wire resistance on performance \cite{10.1145/3505170.3506731, 10.1145/3508352.3561379}. Pin constraint awareness \cite{10.1145/3505170.3506731} is also critical in macro placement to ensure that macros are placed in locations that meet their pin connectivity requirements.

\subsection{Problem Formulation}
\label{sec:problem}
We formulate the macro placement problem as a sequential Markov decision process (MDP) in which an RL agent sequentially places $T$ macros (${\mathcal M}_0$ to ${\mathcal M}_{T - 1}$) onto a layout area or chip canvas (we use both terms). The problem has the following components.
\begin{itemize}

\item \textbf{States:} $s_t$ encodes the observed information collected from the RL environment at the current placement step $t$.

\item \textbf{Position action:} $a^P_t$ is drawn from an action space ${\mathcal A}^P$, represented by $N_{max} \times N_{max}$ discrete actions ($N_{max}$ is set to $128$) that correspond to all possible locations on the chip canvas where the current macro could be placed without overlapping with already placed elements. However, to reduce the search space, the canvas is generally divided into smaller areas of $N_r \times N_c$ grid cells, where $N_r$ are the grid rows and $N_c$ are the grid columns. $N_r$ and $N_c$ should not be greater than $N_{max}$.

\item \textbf{Position mask:} $m^P_t$ is an $N_{max} \times N_{max}$ matrix with 0 or 1 entries that represent which positions are free for placement (value 1) or occupied (value 0). Positions outside the chip canvas are marked (masked) as unplaceable areas.

\item \textbf{Reward:} the $\mathcal R$ is calculated at the end of a placement, meaning once all macros and standard cell clusters have been placed on the chip canvas. It is a negative weighted sum of several proxy costs: wirelength (${\mathcal C}_W$), congestion (${\mathcal C}_C$), and density (${\mathcal C}_D$)

\end{itemize}

The RL agent modeled by the deep neural network is trained to maximize an expected cumulative reward as follows:
\begin{align}
{\mathcal J}(\theta) = \mathbb{E}_{p \sim \pi(\theta)} \Big[ {\mathcal R}_p \Big], 
\label{eq:formula}
\end{align}
where $\theta$ represents the parameters of the RL model, and $p$ represents episodes drawn from the policy distribution $\pi(\theta)$. The placement is constrained by the following four requirements: (1) macros and layout areas can be rectilinear polygons \cite{10.1145/288548.288623, 10.1109/72.623207, 10.5555/244522.244865, 10.1109/43.144848}; (2) macros are placed in groups based on the design hierarchy (\textbf{design hierarchy bias}); (3) macros are restricted to be placed near the periphery (\textbf{peripheral bias}); (4) macros are placed so that pins are not blocked for their connections with other standard cells and macros (\textbf{pin accessibility}).

\section{Methodology}

\begin{figure}[htbp]
\centering
\includegraphics[width=0.9\linewidth]{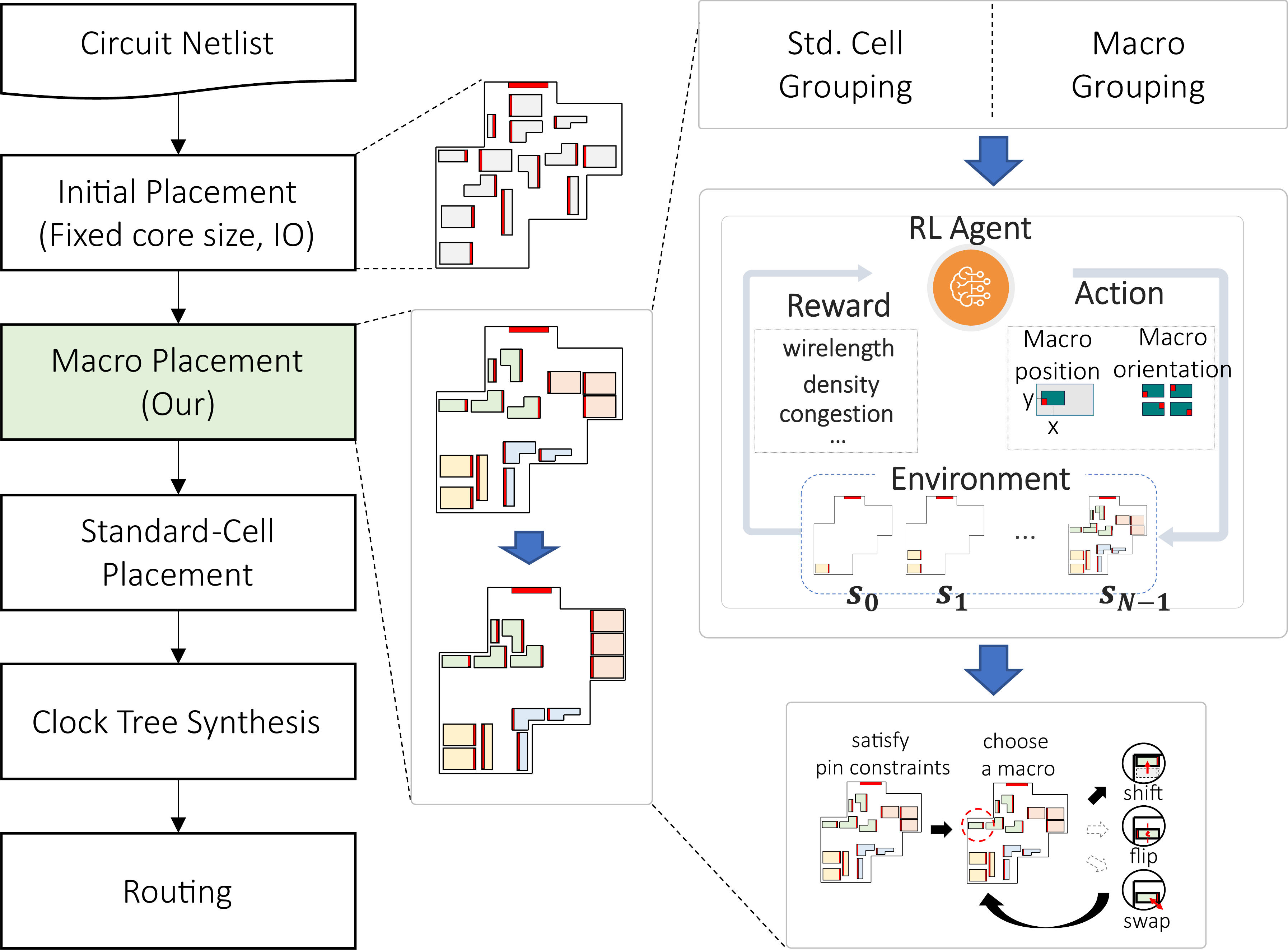}
\caption{Our macro placer in physical design flow.}
\label{fig:flow}
\end{figure}

As shown in Figure \ref{fig:flow}, our framework consists of three distinct engines designed to optimize the processes of standard cell and macro grouping, macro placement, and post-processing placement. First, the grouping engine groups millions of standard cells into several clusters and classifies all the macros into groups based on the design hierarchy. It can be guided by human or automatically inferred from the netlist (as we did not have access to the original RTL). Second, the RL-based placement engine receives input from the grouping engine and produces near-final placements. This engine uses methods to handle rectilinear macros and layout areas, and to satisfy constraints about the design hierarchy, and peripheral bias. Third, the SA-based post-placement engine fine tunes the results generated by the RL placement engine for better pin accessibility, and dead-space minimization.

\subsection{Grouping Engine}
\label{sec:grouping}

\begin{figure}[htbp]
\centering
\includegraphics[width=0.9\linewidth]{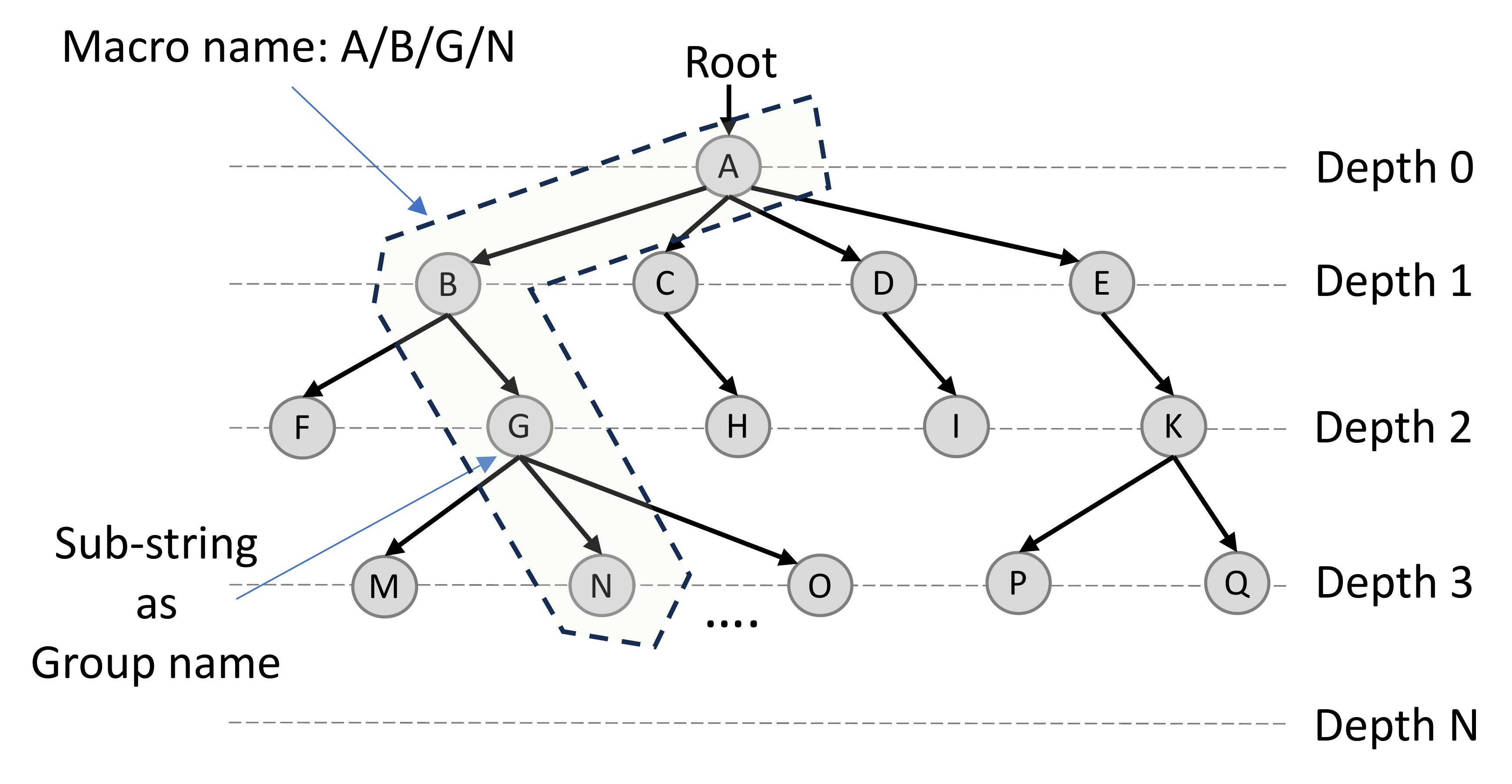}
\caption{An example of tree data structure.}
\label{fig:macro_grouping}
\end{figure}

The grouping of standard cells is done in a similar way to Google method \cite{mirhoseini2021graph}, which utilizes hMETIS \cite{10.1145/266021.266273} as its underlying partitioning algorithm. Our main focus is on grouping macros, where we allow for human guidance. When human guidance is not possible, we propose an alternative method which analyzes the names of all macros in the netlist and identifies the group information by constructing a tree data structure composed of common sub-strings. This is based on the expectation that names will reflect the original hierarchy to some degree. Figure \ref{fig:macro_grouping} illustrates the resulting tree data structure, which consists of nodes that represent sub-strings found in the macro names. To identify proper groups, a recursive search procedure is implemented at each depth level of the tree, such that all nodes in the same depth level are traversed. If a node at a given depth level has more than one child, it is considered a group. Otherwise, the search continues to deeper depth levels. In this way, a list of unique groups are generated and used in the RL placement engine.

\subsection{Deep RL-based Placement Engine}

This engine takes the outputs of the grouping engine to initialize the placement simulator (e.g., environment) in which it performs algorithms to handle rectilinear macros in rectilinear layout areas. 

\subsubsection{Rectilinear Macros and Area Handling}

\begin{figure}[htbp]
\centering
\includegraphics[width=0.9\linewidth]{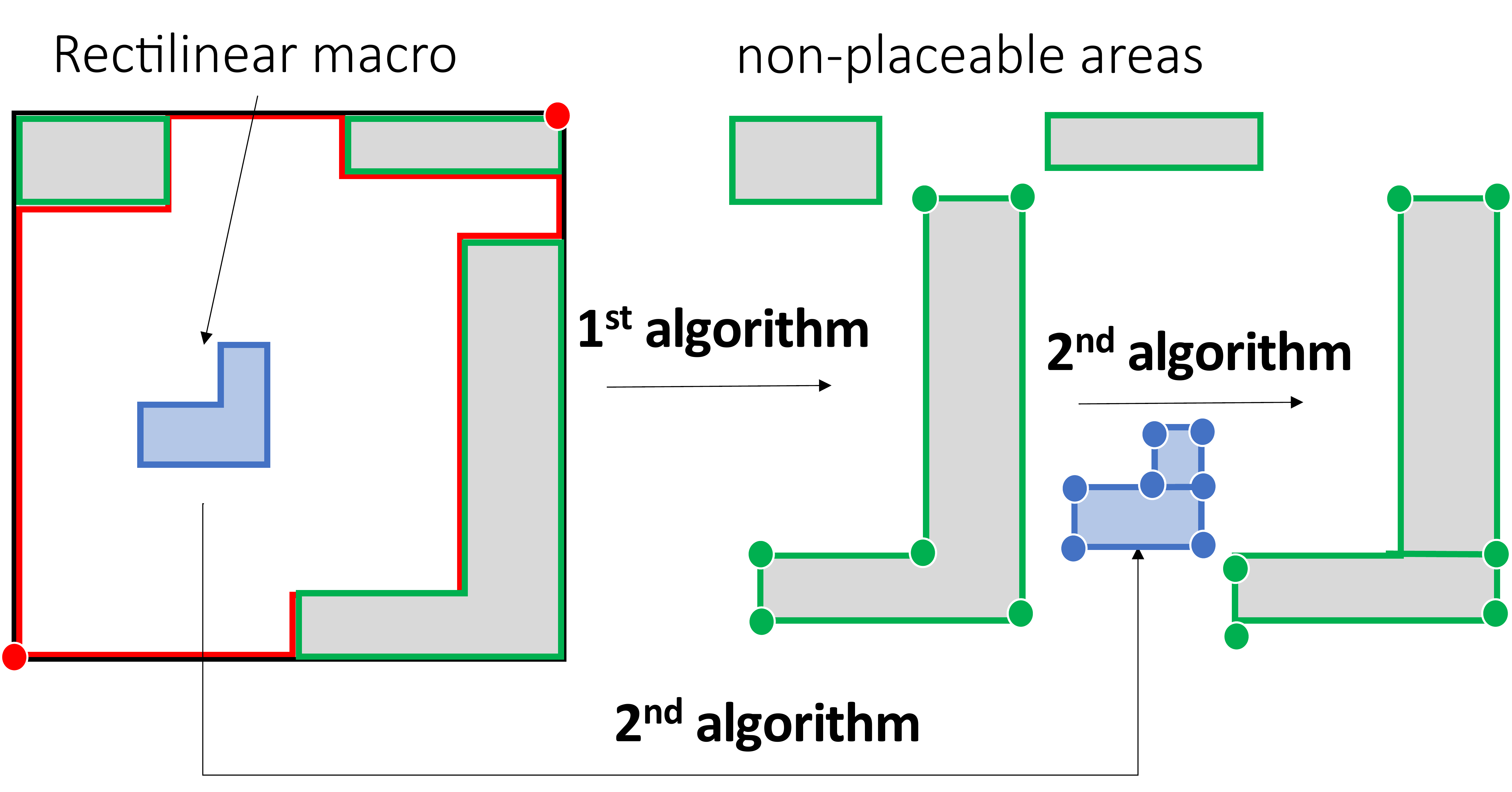}
\caption{Rectilinear macros and layout handling.}
\label{fig:layout_rectangularization}
\end{figure}

We propose two algorithms for handling the placement of rectilinear macros. The first algorithm identifies non-placeable areas, and the second algorithm decomposes each rectilinear shape (non-placeable areas and rectilinear macros) into multiple rectangles as illustrated in Figure \ref{fig:layout_rectangularization}. This representation allows the use of a grid-based masking algorithm (Section \ref{masking}) to work with ``primitive'', i.e. rectangular, blocks and maximize the use of the layout area.

\subsubsection{Masking Control Algorithm}
\label{masking}
\begin{figure}[htbp]
\centering
\includegraphics[width=\linewidth]{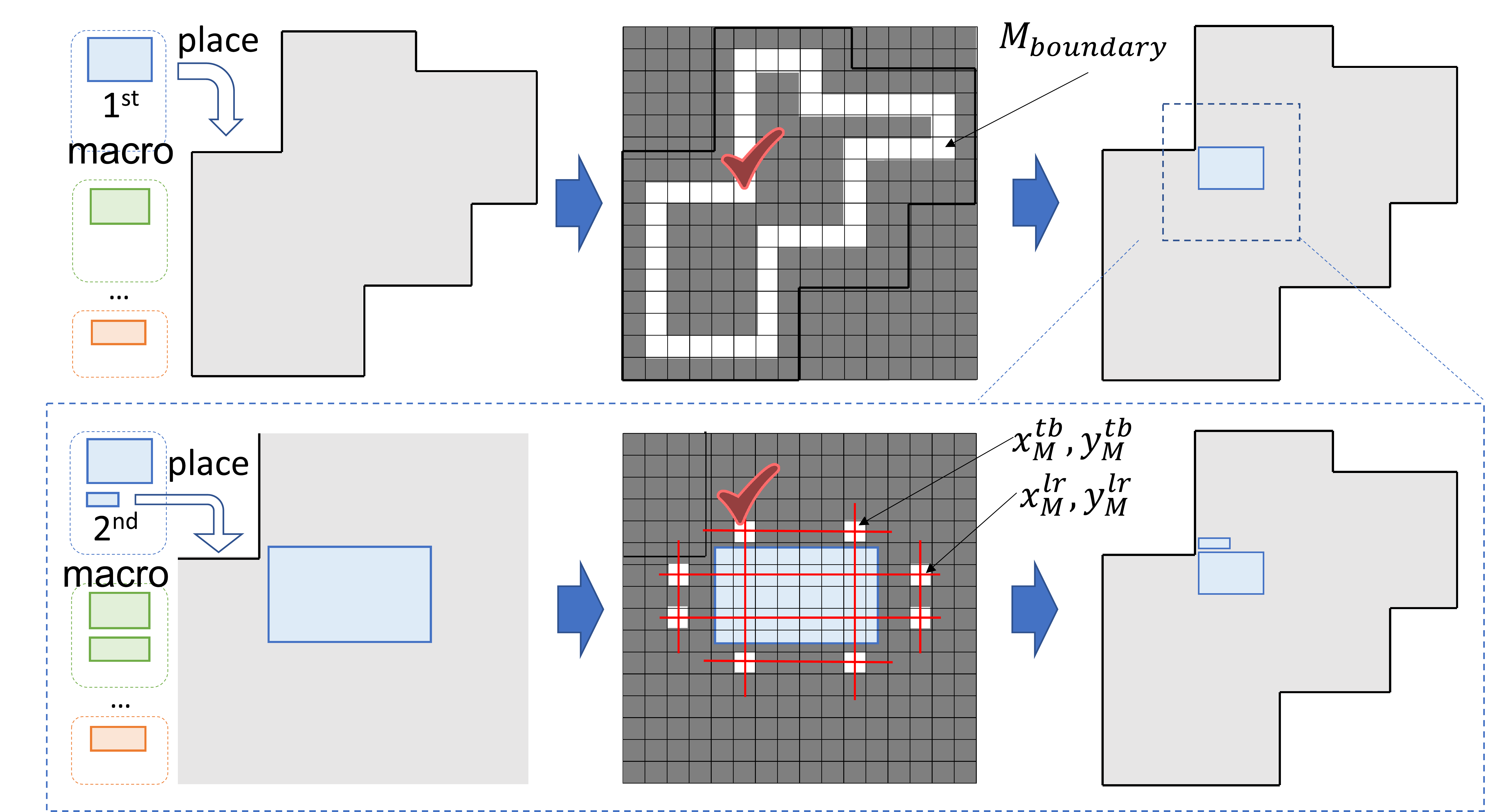}
\caption{An illustration of the masking control algorithm.}
\label{fig:masking}
\end{figure}

Our next step is to control the position mask to ensure that the currently placed macro adheres to the design hierarchy and periphery bias. It is performed repeatedly at beginning of each placement step, and it returns the position mask ($m_j$) of the current macro to be placed (${\mathcal M}^i_j$). As illustrated in Figure \ref{fig:masking}, if the macro is the first from its group, the position mask is the boundary mask ($M_{boundary}$), which allows the macro to be placed only by the closest peripheral grid cells. Beginning with the second macro of a group, to increase pin accessibility and follow the design hierarchy, the algorithm restricts the placeable grid cells to be in close proximity to macros from the same group that have already been placed. To do that, the algorithm loops through all placed macros from that group and selects grid cells with intersecting rows and columns according to the following criteria.

\begin{align}
x_{\mathcal M}^{lr} &= x_{{\mathcal M}_{P}} \pm (\lfloor \frac{w_{{\mathcal M}_{P}}}{2} \rfloor + \lfloor \frac{w_{{\mathcal M}_{C}}}{2} \rfloor) \nonumber\\
y_{\mathcal M}^{lr} &= y_{{\mathcal M}_{P}} \pm \lfloor \frac{h_{{\mathcal M}_{P}} - h_{{\mathcal M}_{C}}}{2} \rfloor  \nonumber \\
x_{\mathcal M}^{tb} &= x_{{\mathcal M}_{P}} \pm \lfloor \frac{w_{{\mathcal M}_{P}} - w_{{\mathcal M}_{C}}}{2} \rfloor  \nonumber \\
y_{\mathcal M}^{tb} &= y_{{\mathcal M}_{P}} \pm (\lfloor \frac{h_{{\mathcal M}_{P}}}{2} \rfloor  + \lfloor \frac{h_{{\mathcal M}_{C}}}{2} \rfloor),
\label{eq:alignment}
\end{align}
where $x$, $y$ are the center coordinates and $w$, $h$ the size of the macros, and ${\mathcal M}_{P}$ and ${\mathcal M}_{C}$ are the placed macros and the current macro, respectively. $\{x_{\mathcal M}^{lr}, y_{\mathcal M}^{lr}\}$ are left and right cells of the placed macros, while $\{x_{\mathcal M}^{tb}, y_{\mathcal M}^{tb}\}$ are top and bottom cells. Intuitively, the selected cells enable the edge of the current macro to align with the edges of already placed macros.

\subsubsection{Neural network model}

\begin{figure*}[htbp]
\centering
\includegraphics[width=0.9\linewidth]{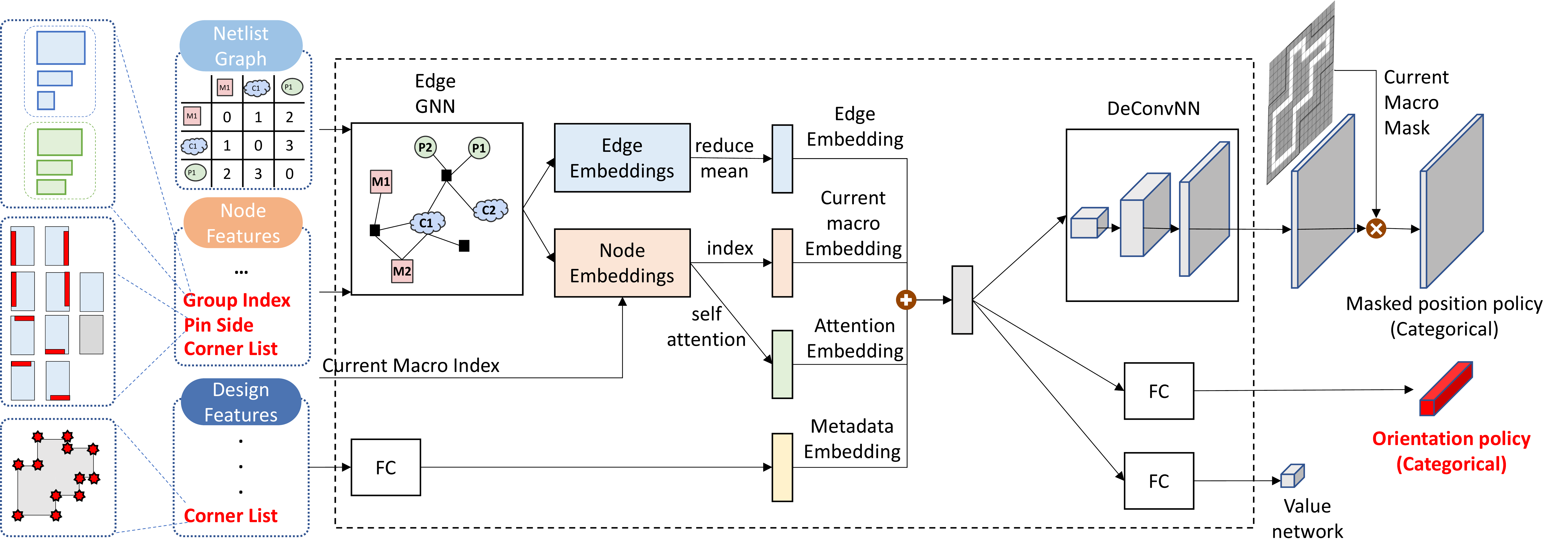}
\caption{Neural network model of RL agent with extra features and policy head for orientation prediction.}
\label{fig:model}
\end{figure*}

Our neural network model (Figure \ref{fig:model}) uses an Edge-GNN to encode the observed information sent from the environment into a low-dimensional vector representation, enabling it to adapt to unseen data. Our proposal makes a key contribution by incorporating additional information that significantly enriches the macro and design features. Specifically, we introduce three crucial elements: the \textbf{group index}, which identifies the group to which a macro belongs; the \textbf{pin side}, which specifies the side of the pin on placed macros; and the \textbf{corner list}, which represents all points in clockwise order from the area of the macro and layout area so that the rectangular and non-rectangular shapes of macros and area can be unified.

Furthermore, as an additional advancement, we upgrade our model to a \textbf{two-head policy} that allows the macro position and orientation to be predicted simultaneously. Actions drawn from the two-head policy create a joint distribution of two categorical distributions that is updated via gradient descent. The PPO\cite{schulman2017proximal} algorithm was chosen to train the RL agent because it is robust to hyperparameters and works well with distributed training systems, allowing to speed up the collection of simulation data.

\subsubsection{Reward function}

Our reward function ${\mathcal R}$ is defined as a negative weighted sum of four proxy costs as follows.

\begin{align}
{\mathcal R} &= - (\alpha {\mathcal C}_W +\beta {\mathcal C}_C +\gamma {\mathcal C}_D +\omega {\mathcal C}_H)
\label{eq:detail_reward}
\end{align}
where ${\mathcal C}_W$, ${\mathcal C}_C$, and ${\mathcal C}_D$ are the three common proxy costs of wirelength, congestion, and density, respectively. The calculation of wirelength cost, density cost, and congestion cost follows known methods (see \cite{mirhoseini2021graph}, \cite{10.1145/3569052.3578926} for example)in which the wirelength cost is approximated as the normalized half-perimeter wirelength (HPWL); the density cost is approximated as the average density of the densest 10$\%$ of grid cells, and the congestion cost is approximated as the average of the top 5$\%$ most congested grid cells. In addition to these conventional proxy costs, we propose a novel proxy cost ${\mathcal C}_H$, which is an extra proxy cost added to the reward function to encourage closeness between macros in the same design hierarchy. It is formulated as follows.
\begin{align}
{\mathcal C}_H = \frac{1}{G}\sum_{g=0}^{G} \frac{\sum_{i,j \in N^g, i \neq j} dist_{ij}}{\sum_{i,j \in N^g, i \neq j} \min {(w_{ij}, h_{ij})}},
\label{eq:hierarchy}
\end{align}

where $G$, and $N^g$ are the number of groups and number of macros in group $g$, respectively. $dist_{ij}$ is the Euclidean distance between two macros in a group and $w_{ij}$, and $h_{ij}$ are the sum of the width and sum of the height of two macros, respectively. Intuitively, the cost will be high if macros of the same group are far apart on the canvas, and the cost will be lower as the macros move toward their expected positions, with each macro close to the other macros from its group.

\subsection{SA-based Post Placement Engine}
To achieve human-quality placement in terms of pin accessibility and dead-space minimization, we propose an SA-based post-placement engine, which incorporates the following key ideas: (1) The engine operates on a dense grid of cells that is linearly scaled with the chip canvas, typically around 2000$\times$2000 cells, providing sufficient precision for macro placement in a reasonable runtime; (2) For each iteration, each macro is applied one of three actions with equal probability: $shift$ (moving a macro to the closest boundary), $swap$ (swapping two macros of the same size, shape, and hierarchy group), or $flip$ (flipping a macro along the x or y axis); (3) If there are any pin accessibility violations present after applying an action, we will revert back to the previous situation.

\section{Experiments}

\textit{Implementation}: We built our framework on top of Google Circuit Training\cite{CircuitTraining2021}. We use \texttt{DreamPLACE}\cite{10.1145/3316781.3317803} to read designs from \texttt{LEF$/$DEF} files and modify it to get the die-area from the macros and layout areas.

\textit{Evaluation designs}: We evaluate the framework using three netlists of Ariane CPU\cite{ariane} provided by \cite{mirhoseini2021graph}, \cite{10.1145/3569052.3578926}, and a version we generated using NanGate45 standard-cell library (NG45). Furthermore, we validated the framework on three industrial designs that contain rectilinear macros and layout areas, implemented originally in an advanced technology node from a commercial foundry. Unfortunately, we are not allowed to disclose exact numbers. Finally, in order to assess the generality of the trained model, we performed training and testing using one Ariane with random rectilinear macros and layouts at NG45 and ASAP7 standard-cell library (ASAP7). Program and evaluation designs are released at \url{https://anonymous.4open.science/r/rl4cad-AE0F}.

\textit{Infrastructure}: Building on top of Google Circuit Training allows our framework to run in distributed fashion across multiple servers and GPUs. However, we constrain our resource utilization to typical configurations. Specifically, our experiments were conducted on a server with a 64-thread CPU, 512 GB of RAM, and an A5000 GPU with 24 GB of memory. Each run uses 25 collectors to gather simulation data.

\textit{Settings}: For comparison purposes, we keep almost all training settings the same as the settings from G-CT\cite{mirhoseini2021graph} and TILOS\cite{10.1145/3569052.3578926}. Additionally, some settings were tuned to work best with our framework: (1) The cost weights $\alpha$, $\beta$, $\gamma$, and $\omega$ in the reward function\footnote{The weights $\alpha$, $\beta$, $\gamma$ in CT weighted cost are 1.0, 1.0, 0.5, respectively, and 1.0, 0.5 and 0.5 in TILOS weighted cost} were set to 5.0, 1.0, 0.5, and 0.1, respectively. Wirelength weight is set to 5.0 to increase the proportion of wirelength proxy which is hard to optimize under the periphery bias constraint; (2) We select the grid size ($N_r$ and $N_c$) relative to the chip canvas so that the smallest macro can fit inside a grid cell; (3) The number of nodes and edges in the RL model is chosen relative to the netlist size (e.g., macros, clusters), and does not cause issues during model updates using GPUs.

\subsection{Evaluations Using the Netlists of Ariane}

\begin{figure*}[htbp]
\centering
\includegraphics[width=0.9\linewidth]{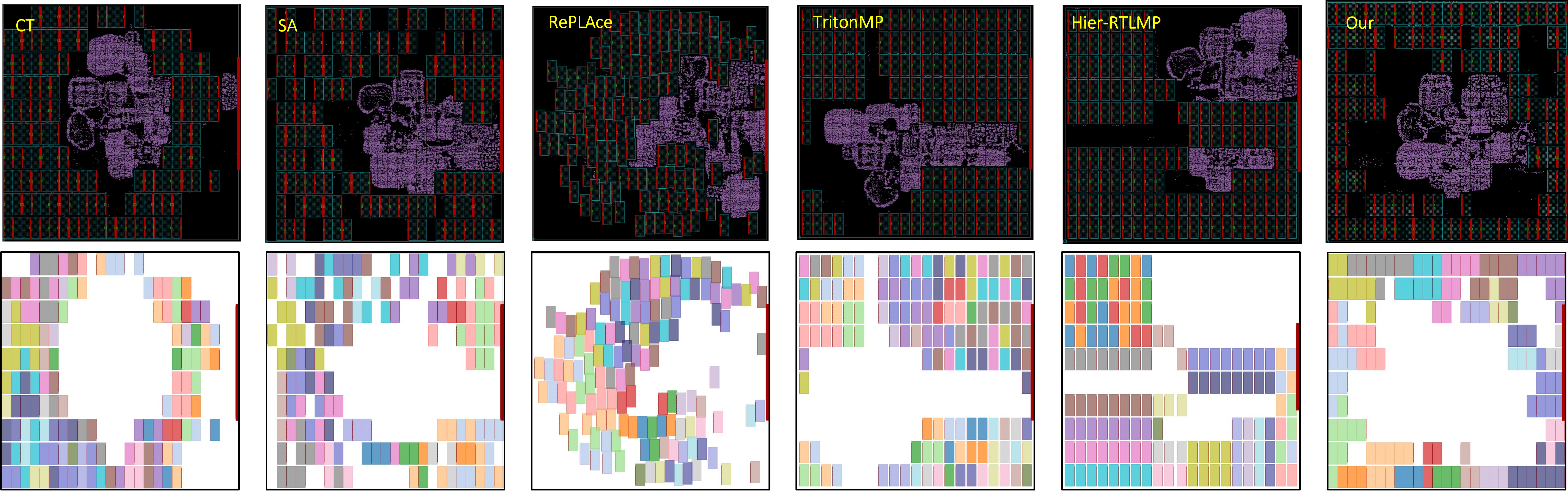}
\caption{Placements on Ariane-NG45 (freq.=770 MHz, density = 64\%) using different academic placers.}
\label{fig:ariane}
\end{figure*}

Our framework can be applied to designs with rectangular macros and areas without losing generality. Table \ref{tab:ariane} describes three different netlists for the Ariane CPU (A-GCT, A-TILOS, and A-OURS) and it shows our results for CT Metrics (block placement with proxy costs averaged over nine runs) and the metrics after we complete the layouts (post block placement) with a commercial P\&R tool (P\&R Metrics, post-route). For the block placement metrics, we compare our results with CT and SA results as published in \cite{CircuitTraining2021, 10.1145/3569052.3578926} as well as our own configuration adding the hierarchy cost. Our method can produce placements that show better proxy cost than those published in \cite{CircuitTraining2021} and \cite{10.1145/3569052.3578926} for both the original configuration (with 8\% and 16.7\% improvement on A-GCT and A-TILOS, respectively) and our configuration (with 8.6\% and 12\% improvement on A-GCT and A-TILOS, respectively).

For the P\&R Metrics, we report results for our netlist\footnote{All metrics from the P\&R Tool are reported after the post-route optimization stage without cleaning DRC errors.} A-OURS, along with the results using CT, SA, and the block placement stage for three academic placers integrated in OpenROAD\cite{OpenROAD}: RePLAce, TritonMP and Hier-RTLMP. As indicated, we complete and measure post-route  with the same commercial tool for consistency. In three out of four metrics, our framework  has the best or second-best results compared to other placers. Figure \ref{fig:ariane} shows physical placements from the systems we studied. Our placer shows similarities to Hier-RTLMP in term of placing macros based on the design hierarchy, as well as similarities to both Hier-RTLMP and TritonMP in placing macros on the periphery. We do not force macros away from the IO ports and leave that human-like rule for further improvements.

\begin{table}[!htbp]
\centering
\caption{Results on different Ariane netlists}\label{tab:ariane}
\resizebox{.48\textwidth}{!}{
\begin{tabular}{|c|c|c|c|c|c|c|c|c|}\hline
\multirow{2}{*}{{Designs}} & \multicolumn{4}{c|}{\textbf{Netlist information}} & \multicolumn{4}{c|}{\textbf{Model Configuration}} \\
\cline{2-9}
                                & {Core}& {$\#$}    & {$\#$}   & {$\#$}     &  {{Ori.}} &  {{Our}}   & {{$\#$}} & {{$\#$}}\\
                                & {Size}& {Macros}  & {IOs}    & {Clusters} &  {{Grid}} &  {{Grid}}  & {{Nodes}} & {{Edges}} \\
\hline
{Ariane} &	356.592 & \multirow{2}{*}{133} &  \multirow{2}{*}{1231}  &  \multirow{2}{*}{799}  &  \multirow{2}{*}{35x33} & \multirow{2}{*}{12x18} &  \multirow{2}{*}{1200} &  \multirow{2}{*}{10000} \\
{(GCT)}   & 356.640 &                       &                         &                        &        &        &  &\\
\hline
{Ariane} &	1347.1 &\multirow{2}{*}{133} &  \multirow{2}{*}{495}   &  \multirow{2}{*}{810}  &  \multirow{2}{*}{23x28} &  \multirow{2}{*}{23x10}                &  \multirow{2}{*}{1200} &  \multirow{2}{*}{12000} \\
{(TILOS)} & 1346.8 &                       &                         &                        &        &                       & & \\
\hline
{Ariane} &	1445.9 & \multirow{2}{*}{133} &  \multirow{2}{*}{495}  &  \multirow{2}{*}{41} &   \multirow{2}{*}{-} & \multirow{2}{*}{25x10} &  \multirow{2}{*}{200} &  \multirow{2}{*}{1100} \\
{(OURS)}  & 1444.8 &                       & &                        &                         &    & & \\
\hline
\multicolumn{9}{|c|}{\textbf{CT metrics}}\\\hline
\multirow{2}{*}{{Designs}} & {{Placer}} & {WL}   & {Den.} & {Cont.} & {Hier.}  & {CT} & {Our} & {Inference}\\
                                & {} & {Cost} & {Cost} & {Cost}  & {Cost} & {Cost} & {Cost} & {time(h)} \\\hline
\multirow{5}{*}{Ariane} &	{CT\cite{CircuitTraining2021}} &  0.1013  &  0.5502  &  0.9174  & - & 1.1102 & - & -\\
\cdashline{2-9}
\multirow{5}{*}{(GCT)}&	{CT$_{(12\times18)}$} &  \textbf{0.0886}  &  0.5345  &  0.8852  & 2.2115  & - &  1.6411 & 0.02 \\

& {SA$_{(12\times18)}$}      & 0.0963 & \textbf{0.5057}   & 0.8446   & 1.4281  & - & 1.5523 & 14 \\

& {Our$_{RL}$}   & 0.0973 & 0.5088 & 0.8507 & 1.0571 & 1.0315 & 1.5264 & 0.02 \\

& {Our$_{POST}$}    & 0.0933  & 0.5070  & \textbf{0.8414}  & \textbf{1.0565}  & \textbf{1.0209} & \textbf{1.4997} & 0.1 \\
\hline

\multirow{5}{*}{Ariane}  & {CT\cite{10.1145/3569052.3578926}} & 0.1060 & 0.5280 & 1.0470 & -  & 0.8932 &- & - \\

\multirow{5}{*}{(TILOS)}& {SA\cite{10.1145/3569052.3578926}} & 0.0860  &0.4990 & 0.8350 & -  & 0.7533 & - & 12.5 \\
\cdashline{2-9}
& {CT$_{(23\times10)}$}     & \textbf{0.0975}  & 0.5860  & 0.7881 & 2.9580 & - & 1.7635 & 0.02 \\

& {SA$_{(23\times10)}$} & 0.1061 & \textbf{0.5038} & 0.7761 & 1.5988 & - & 1.5820 & 10 \\

& {{Our$_{RL}$}} & 0.1092 & 0.5121 & 0.7701 & \textbf{1.3207} & 0.7503 & 1.5752 & 0.02 \\

& {Our$_{POST}$} & 0.1045 &  0.5156 & \textbf{0.7643} & 1.3211 & \textbf{0.7444} &  \textbf{1.5522} & 0.1 \\
\hline

\multicolumn{9}{|c|}{\textbf{P\&R Metrics (post-route)}} \\
\hline

\multirow{2}{*}{{Designs}} & {{Placer}} & {Area}       & {WNS}  & {TNS}   & {$\#$} & {Power} & {Proxy} & {Inference}\\
                                & {} & {(mm$^2$)} & {(ns)} & {(ns)}  & {DRC}  & {(mW)} & {cost} & {time (h)} \\\hline
\multirow{5}{*}{Ariane} & {CT$_{(25\times10)}$} & 1.2806 & -0.91 & -4833.9 & 9 & 585 & 1.8570 & 0.02\\
\multirow{5}{*}{(OURS)} & {SA$_{(25\times10)}$} & 1.2850 & -0.93 & -5320.6 & 9 & 586 & 1.7879 & 14\\
& {RePLAce} & 1.2812 & -1.04 & -5423.7  & 9 & 584 & 1.7244 & 1\\
& {TritonMP} & 1.2839 & -0.89 & -5068.2  & 9 & 586 & 1.9621 & 1\\
& {Hier-RTLMP} & 1.2823 & -0.84 & -4632.2 & 7 & 586 & 1.6482 & 8\\
& {Our} & 1.2803 & -0.86 & -4731.0 & 6 & 586 & 1.5807 & 0.1\\
\hline
\end{tabular}
}
\end{table}

\textit{Discussions}: Here we propose explanations for the results: (1) Our masking strategy and model enhancement show advantages in reserving as much space as possible for standard cells in the middle area. This brings a reduction in density and congestion costs, and consequently, the total weighted cost. (2) Our placer  finds placements that better preserve the design hierarchy, leading to a better hierarchy cost and thus the total weighted cost. (3) Our strategy for choosing grid size results in a significant reduction in density and congestion when compared to  previously reported work.


\subsection{Evaluation of Industrial Designs}
For each design, we selected the top three placements from the training phase and evaluated them using a commercial, state-of-the-art P\&R tool. The best results are reported in Table 2 together with results from the timing-driven placer from the commercial P\&R tool (Comm) and from the same tool but aided by designers (Human). We only applied reasonable efforts (no ``benchmarking'') , meaning we wanted to see if results were comparable, and not to try to prove if any such approach could ``beat'' the others. The netlists cannot be disclosed. 
Our placer achieved PPA results that are better than those obtained by the designers within a few evaluations and are quite comparable to those achieved by the timing-driven placer from the P\&R tool. The output from our placement is entered into the commercial tool without any additional modification (See Fig \ref{fig:flow}) nor prior knowledge about the power grid or later steps. This may create some DRC errors. Improving this aspect is a long-term plan for our future work.

\begin{table}[!htbp]
\centering
\caption{Results of industrial designs}\label{tab:industry}
\resizebox{.48\textwidth}{!}{
\begin{tabular}{|c|c|c|c|c|c|c|c|}\hline
\multirow{2}{*}{{Designs}} & {$\#$} & {$\#$}   & {$\#$}   & {{$\#$}}     &  {{$\#$}} &  {{Recti.}}   & {{Recti.}}\\
                                & Macro & Types & {IOs}    & {Cells} &  {{Nets}} &  {{Layout}}  & {{Macros}} \\
\hline
\multicolumn{1}{|c|}{ic1} &	89 & 59  &  1125  &  1.5M  &  1.7M  & \checkmark &   \\
\hline
\multicolumn{1}{|c|}{ic2} &	169 &97  &  630  &  3.8M  &  4.3M & \checkmark &   \\
\hline
\multicolumn{1}{|c|}{ic3} &	94 & 21  &  2207  &  1.8M  &  1.8M & \checkmark &  \checkmark \\
\hline

\multicolumn{8}{|c|}{\textbf{Layout Metrics}}\\\hline
\multirow{2}{*}{{Designs}} & \multirow{2}{*}{Placer} 
& Area   & WNS & TNS & $\#$ & Power & Run \\
&& (mm$^2$) & (ns) & (ns)  & DRC  & (mW)  & time(h) \\
\hline

\multirow{3}{*}{ic1} 
& Human      & 0.4550 & -0.6201 & -0.6201 & 2559 & 44.6 & weeks\\
& Comm       & \textbf{0.4495} & \textbf{-0.6044} & \textbf{-0.6044} & \textbf{2491} & 46.8 & 0.5 \\
& Our    	 & 0.4548 & -0.6178 & -0.6178 & 2695 & \textbf{43.7} & 14\\
\hline

\multirow{3}{*}{ic2} 
& Human      & 1.0331 & -0.0709 & -376.68 & \textbf{6619}  & 62.6 & weeks\\
& Comm 		 & 1.0256 & -0.0739 & -302.11 & 23088 & \textbf{58.5} & 12 \\
& Our        & \textbf{1.0206} & \textbf{-0.0698} & \textbf{-288.59} & 23542 & 59.8 & 28\\
\hline

\multirow{3}{*}{ic3} 
& Human      & 5.7972 & -0.4193 & -1.4651 & \textbf{3924} & 284 & weeks\\
& Comm 		 & 5.7965 & -0.4544 & -15.5075 & 5038 & 274 & 1.7 \\
& Our        & \textbf{5.7961} & \textbf{-0.1402} & \textbf{-0.5792} & 4313 & \textbf{269} & 14\\
\hline
\end{tabular}
}
\end{table}

\subsection{Generalization between Technology Nodes}

\begin{figure}[htbp]
\centering
\includegraphics[width=\linewidth]{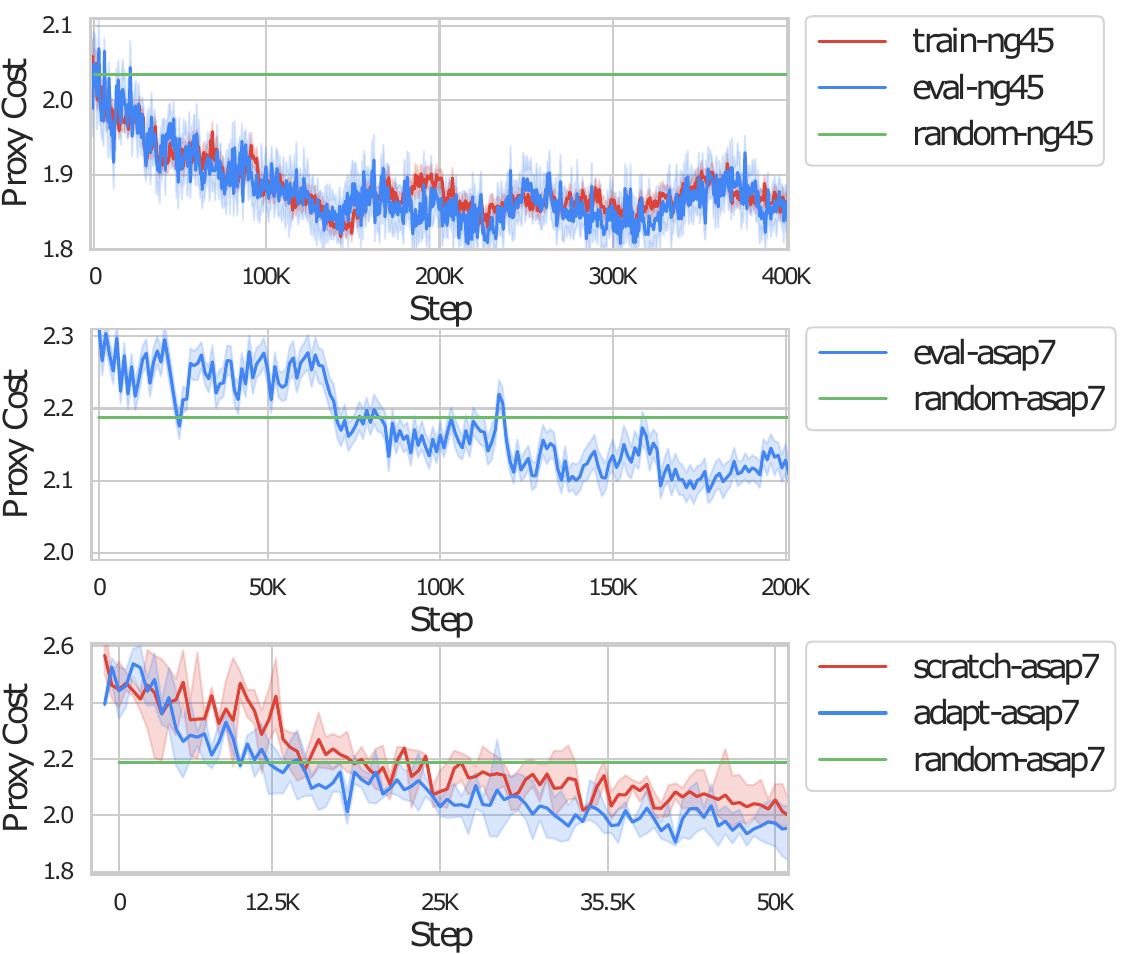}
\caption{Evaluation curves on Ariane-NG45 (top), Ariane-ASAP7 (middle), and adaptation to a harder Ariane-ASAP7.}
\label{fig:generalize}
\end{figure}

\begin{figure}[htbp]
\centering
\includegraphics[width=\linewidth]{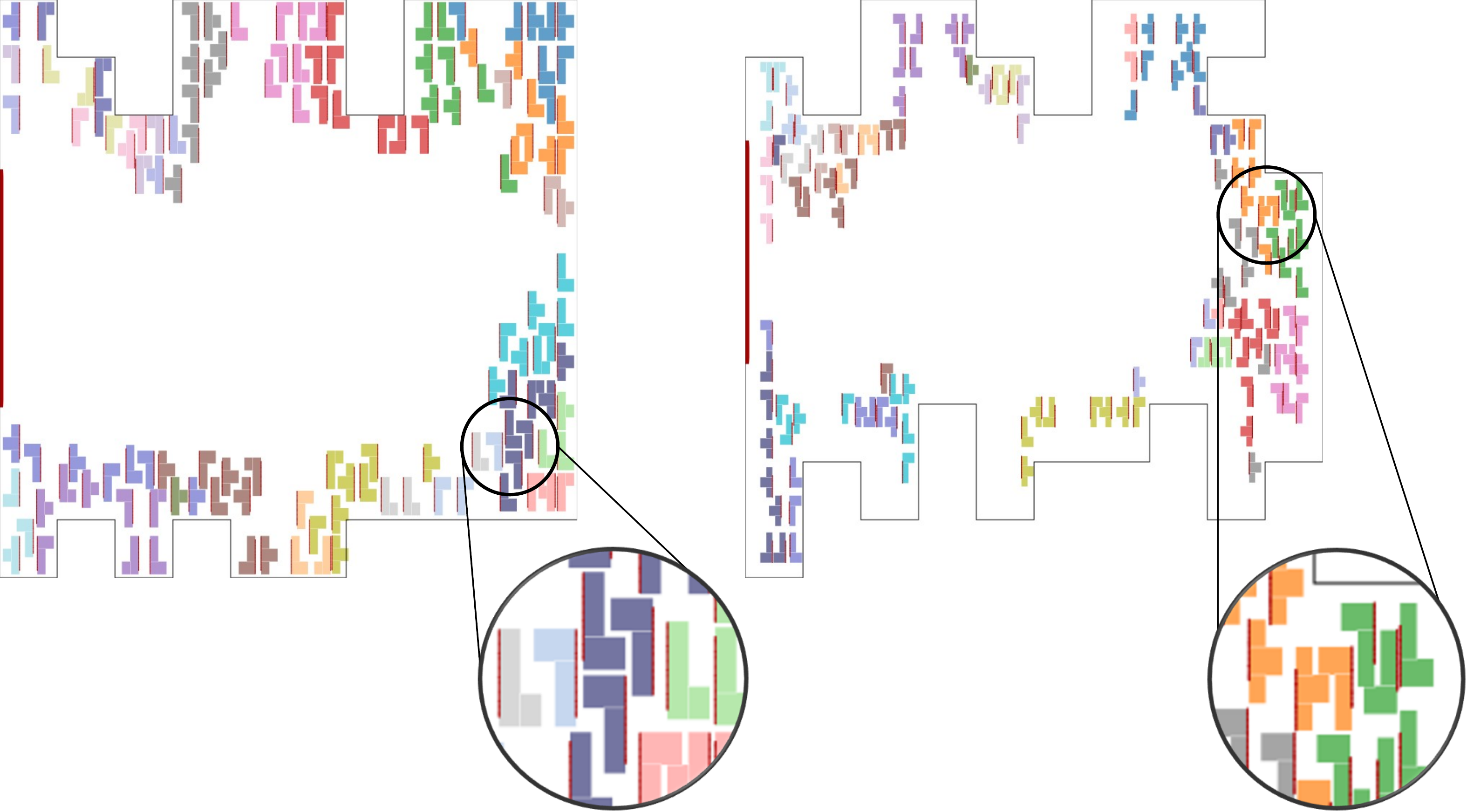}
\caption{Post-placement on Ariane (NG45 and ASAP7).}
\label{fig:visual_image}
\end{figure}

The last experiment assessed the possible generalization of our model to designs containing rectilinear macros and areas. In this study, we restricted the macro shapes to L, J and T patterns\cite{10.1109/ASPDAC.1997.600145}, and avoided modifying macro shapes on their IO sides. The top plot from Figure \ref{fig:generalize} shows the training and evaluation curves from 80 synthesized designs used for training and 20 synthesized designs used for testing. The model is well trained to handle rectilinear designs from NG45. When we evaluated with 100 synthesized designs at ASAP7, the model-generated placements improved outperforming the random placements after 100K policy updates (middle plot of Figure \ref{fig:generalize}). Finally, we tested the transferability of the best model trained on NG45 to a more challenging synthesized design at ASAP7. The bottom plot of Figure \ref{fig:generalize} shows that adapting from a pre-trained model enabled the model to converge faster than training the model on that design from scratch. Figure \ref{fig:visual_image} depicts placements on unseen designs of Ariane at NG45 and ASAP7, which are generated by the trained model and are fine-tuned with SA-based post-placement.

\subsection{Runtime analysis}

Table \ref{tab:ariane} and Table \ref{tab:industry} reports the inference time and training time of various placers. Our learning-based placer only needs a few minutes to obtain a good placement. However, to generate a well-trained agent, the agent needs a few hours of training. Specifically, 14 hours are needed with the G-CT netlist and 10 hours with the TILOS netlist, which increases the total runtime. It's worth noting that with the same amount of training time, the reported work required a farm of CPUs and GPUs (20$\times$96vCPUs and 8$\times$V100s), while our placer has been optimized to consume minimal computing resources and conform to common EDA server configurations. Using an adaptation technique from a pre-trained agent is a way to reduce the training time to 6 hours\cite{mirhoseini2021graph}. However, keeping the runtime of a RL-based placer to be the same as that of traditional placers is a challenging problem. We will address this in our future work.

\section{Conclusion}

This work proposed an RL-based macro placement framework that exceeds the performance of reported work and achieves comparable results to existing baseline algorithms. It strictly respects crucial human-like constraints, with a specific focus on design hierarchy and peripheral bias. Furthermore, this approach has the potential to generalize a learned model to various designs with rectilinear macros and areas. Lastly, our advances, conducted on standard training machines, can drive the research in RL-based placement towards efficiency and affordability allowing IC design house to adopt it without adding excessive computing resources.

\begin{acks}
We thank Prof. A. Kahng from UC San Diego  for sharing the benchmark data, and Dr. A. Domic from Kepler Computing for his useful feedback. This work was supported by the Korean TIPA R\&D Program (S3207298)
\end{acks}
\balance
\bibliographystyle{ACM-Reference-Format}
\bibliography{references}

\clearpage

\end{document}